\documentclass{article}

\usepackage[preprint]{neurips_2026}

\usepackage[utf8]{inputenc} 
\usepackage[T1]{fontenc}    
\usepackage{hyperref}       
\usepackage{url}            
\usepackage{booktabs}       
\usepackage{amsfonts}       
\usepackage{nicefrac}       
\usepackage{microtype}      
\usepackage{xcolor}         

\usepackage{pgfplots}      
\pgfplotsset{compat=1.18}

\providecommand{\ie}{\textit{i.e.}}
\usepackage{graphicx}
\usepackage{amsmath}
\usepackage{amssymb}
\usepackage{booktabs}
\usepackage{makecell} 
\usepackage{siunitx} 
\usepackage{multirow}
\usepackage{afterpage}
\usepackage{tabularx}

\title{Learning Stable Canonical Worlds for Novel View Synthesis and Beyond}

\author{%
Xiaoyu Xu$^{1}$,
Jian Zou$^{1}$,
Sheyang Tang$^{2}$,
Zhihua Wang$^{1}$,
Jing Liao$^{1}$,
and Kede Ma$^{1}$\thanks{Corresponding author.} \\
$^{1}$Department of Computer Science, City University of Hong Kong \\
$^{2}$Department of Electrical and Computer Engineering, University of Waterloo \\
Emails: \texttt{xiaoyxu@cityu.edu.hk},
\texttt{jian.zou@my.cityu.edu.hk},\\
\texttt{sheyang.tang@uwaterloo.ca}, 
\texttt{Zhihua.WANG@cityu.edu.hk},\\
\texttt{jingliao@cityu.edu.hk},
\texttt{kede.ma@cityu.edu.hk}
}

\begin{document}
\maketitle

\begin{abstract}
Feed-forward Gaussian splatting (FFGS) facilitates real-time novel view synthesis, yet current methods often remain tied to view-dependent predictions. As more input views are added, they may accumulate noisy or redundant evidence instead of converging to a stable scene representation. In this paper, we introduce CanonicalGS, a feed-forward pipeline that maps cluttered multi-view observations into a stable, scene-centric representation. CanonicalGS first extracts view-centric evidence from depth, semantic features, and uncertainty estimates, and then aggregates this evidence in a canonical latent world using uncertainty-aware fusion. By emphasizing reliable observations while suppressing uncertain or redundant ones, CanonicalGS produces representations that scale more effectively for novel view synthesis and transfer to downstream visual perception tasks. Experiments show up to a $2.5$ dB improvement in peak signal-to-noise ratio for synthesizing novel views and an $11\%$ gain in semantic segmentation accuracy.
\end{abstract}
    
\section{Introduction}
\label{chapter:introduction}

Human vision does not treat a scene as a loose collection of independent views. Instead, it integrates partial, viewpoint-dependent observations into a stable internal model of the physical world~\citep{Marr1978RepresentationAR}. Marr's theory of vision offers a useful computational account of this process: visual perception progresses from image measurements, to intrinsic 2.5D descriptions such as depth and surface orientation, and finally to canonical 3D scene representations~\citep{Marr1983VisionAC,Barrow1978RECOVERINGIS,Ullman1996HighLevelVO}. We refer to this transformation from cluttered, view-centric observations into a coherent scene-centric representation as \emph{canonization}. Although this idea has shaped vision theory for decades, realizing it in learnable 3D computational systems has remained difficult, partly because earlier scene representations lacked the flexibility and expressiveness needed for end-to-end visual learning and understanding.

Recent advances in 3D representation learning make this classical objective newly practical. Specifically, 3D Gaussian splatting (3DGS) provides an explicit and efficient representation for high-quality scene rendering~\citep{kerbl20233d,fu2024colmap,charatan2024pixelsplat}. Building on it, feed-forward Gaussian splatting (FFGS) methods~\citep{charatan2024pixelsplat,chen2024mvsplat,xu2025depthsplat,liu2025monosplat} directly infer 3D Gaussian primitives (GPs) from sparse input images in a single forward pass. Viewed through Marr's representational hierarchy, this computation has the ingredients for canonization: images are lifted through geometric cues and expressed as 3D primitives. The difficulty is that this promise is usually pursued through a rendering-first objective, where multi-view evidence is optimized to synthesize target views rather than to form a shared scene representation.

This rendering-first bias appears across existing FFGS methods. Pose-required approaches often build on view-dependent or pixel-aligned predictions~\citep{charatan2024pixelsplat,chen2024mvsplat,xu2025depthsplat,liu2025monosplat}, while pose-free and aggregation-based variants improve input flexibility and cross-view fusion~\citep{ye2024no,huang2025no,ye2025yonosplat,kang2025selfsplat,zhang2025flare,jiang2025anysplat,fei2024pixelgaussian,li2026tokensplat,wang2025volsplat,itkin2026globalsplat,miao2025evolsplat}. However, these methods still optimize primarily for renderable reconstruction, so noisy or redundant observations can be passed to the resulting GPs rather than resolved in a shared scene representation as the input set grows. This motivates a computational pipeline that first consolidates multi-view evidence in (latent) scene space and then decodes the resulting scene representation into GPs for rendering.

In this work, we propose \textbf{CanonicalGS}, a canonization-oriented FFGS method that reconnects Marr's classical representational progression with modern 3D Gaussian representations. The core idea is to convert noisy and partial view-centric evidence into a canonical latent world that preserves reliable and stable scene information while suppressing uncertainty and redundancy. Concretely, CanonicalGS follows a 2D$\rightarrow$2.5D$\rightarrow$3D pipeline. Given a set of input images, our method first estimates 2.5D depth maps that provide both geometric structure and reliability cues. Next, it aggregates this evidence in scene space, constructing reliability and feature fields in world coordinates so that reliable observations are consolidated while uncertain or inconsistent ones are suppressed. A GP decoder then maps the latent scene representation into a renderable Gaussian field, encouraging each primitive's contribution to increase or saturate with accumulated reliability rather than decrease as reliable evidence is added.

CanonicalGS changes the role of FFGS from a computational pipeline for novel view synthesis into a representation learning algorithm for broader visual perception and understanding. Such downstream capability is not imposed through auxiliary perception heads or external semantic guidance. Instead, it emerges from the architectural pressure to consolidate reliable evidence into a shared scene space before decoding. Extensive experiments show that CanonicalGS improves rendering under increasing input views and yields more transferable scene representations: it achieves up to a $2.5$ dB improvement in peak signal-to-noise ratio (PSNR) for novel view synthesis and an $11\%$ gain in semantic segmentation accuracy.
\section{Related Work}
\label{chapter:related_works}

We position CanonicalGS within three related lines of work: FFGS, multi-view aggregation, and uncertainty-aware reconstruction. We focus on how each line handles scene representation, evidence consolidation, and reliability, which together motivate our canonization-oriented design.

\textbf{FFGS.} 3DGS represents scenes with explicit GPs and has become an efficient alternative to neural radiance-field rendering~\citep{kerbl20233d}. Subsequent work has improved the flexibility and expressiveness of 3DGS through alternative primitive formulations and factorized appearance models~\citep{hamdi_2024_CVPR,tang20243igs}. FFGS builds on it by predicting GPs in a single forward pass, avoiding costly per-scene optimization~\citep{charatan2024pixelsplat,szymanowicz2024splatter}. Pose-required FFGS methods use known or estimated cameras together with geometric priors, cost volumes, monocular depth, or multi-view cues to infer GPs from sparse posed views~\citep{charatan2024pixelsplat,fei2024pixelgaussian,chen2024mvsplat,liu2025monosplat,xu2025depthsplat}. Pose-free variants further remove the need for externally provided camera poses by jointly estimating camera relations, geometry, and GPs from unposed images~\citep{ye2024no,huang2025no,kang2025selfsplat,zhang2025flare,ye2025yonosplat}. These methods have substantially advanced fast view synthesis, but they remain primarily optimized around renderable reconstruction. CanonicalGS instead treats FFGS as a representation learning problem: multi-view inputs are first organized into a shared scene space, and GPs are decoded only after scene-centric evidence has been consolidated.

\textbf{Multi-view aggregation.} As FFGS systems move beyond fixed two-view or sparse-view settings, cross-view aggregation becomes central to eliminating redundancy, resolving inconsistent observations, and supporting larger input sets. Recent methods explore aggregation in several spaces, with an important distinction between aggregation before and after GP decoding. PixelGaussian and TokenSplat align or aggregate image-level evidence before GP prediction~\citep{fei2024pixelgaussian,li2026tokensplat}; GlobalSplat summarizes scenes through compact global tokens~\citep{itkin2026globalsplat}; and VolSplat and EvolSplat introduce voxel- or volume-aligned representations before decoding GPs~\citep{wang2025volsplat,miao2025evolsplat}. AnySplat instead merges or routes information in Gaussian space after primitive hypotheses are formed~\citep{jiang2025anysplat}. Together, these approaches show that FFGS benefits from separating evidence consolidation from direct primitive prediction. However, aggregation alone does not determine whether additional views shall refine the scene representation or simply inject more view-conditioned hypotheses. CanonicalGS is closest in spirit to methods that aggregate before decoding, especially voxel-aligned approaches, but differs by making evidence accumulation uncertainty-aware and by further conditioning the GP decoder to accumulated reliability, so that reliable observations are encouraged to strengthen rather than destabilize the representation.

\textbf{Uncertainty-aware reconstruction.}
Classical multi-view reconstruction has long relied on photometric consistency, visibility reasoning, and confidence or filtering cues to reject ambiguous matches, suppress unreliable depth estimates, and fuse observations robustly under occlusion and noise~\citep{curless1996volumetric,seitz2006comparison,furukawa2010accurate}. Modern learning-based multi-view stereo derives uncertainty or confidence from depth distributions, matching costs, attention, or cross-view consistency, and uses it to refine or filter local geometry~\citep{yao2018mvsnet,luo2020attention,cheng2020deep,yang2022non}. Related ideas also appear in dense reconstruction, where probabilistic or learned reliability measures guide the integration of multiple noisy measurements into a coherent 3D estimate~\citep{weder2021neuralfusion,rosinol2023probabilistic}.
These studies establish an important principle: not all observations should contribute equally to scene construction. 
CanonicalGS brings this principle into FFGS by using uncertainty not merely as an auxiliary depth cue, but as a first-class aggregation signal that weights reliable geometric and appearance evidence before GP decoding.
\begin{figure*}[t!]
	\centering
	\includegraphics[width=1.0\linewidth]{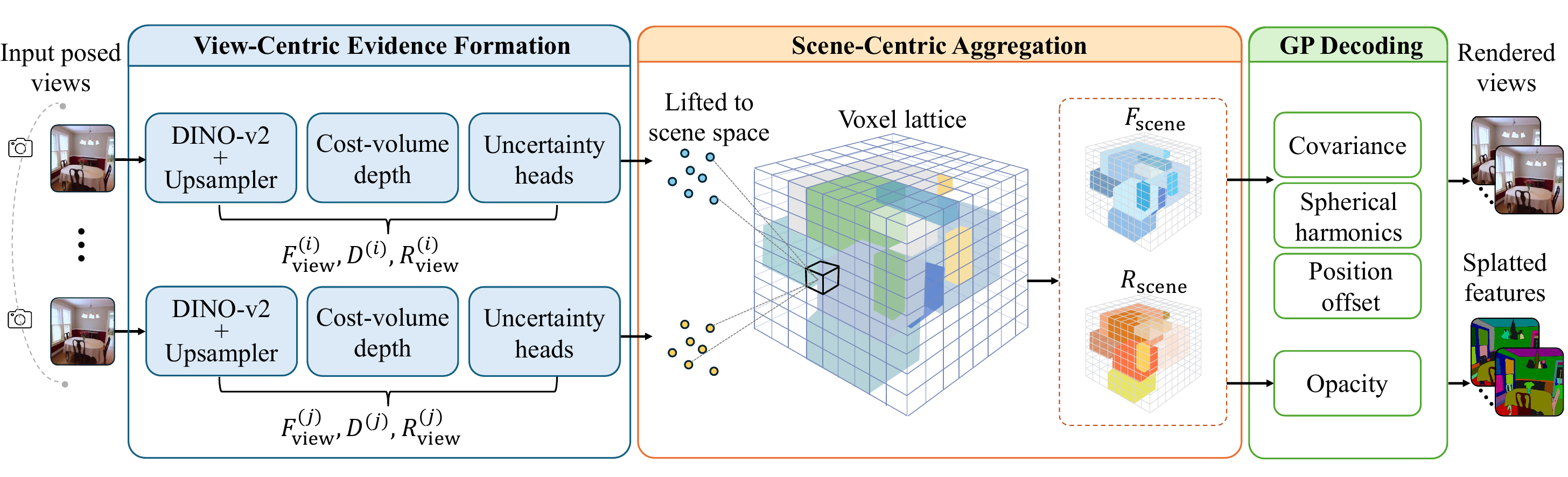}
    \caption{\small Overview of CanonicalGS. CanonicalGS converts posed input views into view-centric evidence, aggregates reliable observations in a scene-centric latent representation, and decodes the resulting scene fields into GPs for novel view synthesis and downstream perception. Redder colors indicate higher reliability in $R_{\mathrm{scene}}$. The illustration shows two views for clarity, but the formulation applies to an unordered set of $N$ views.}
	\label{fig:framework}
\end{figure*}

\section{Proposed Method: CanonicalGS}
\vspace{-1mm}
\label{chapter:method}

In this section, we present CanonicalGS, a representation-first FFGS method that constructs a canonical latent world before decoding GPs for rendering.
\subsection{Problem Formulation}
\vspace{-1mm}

Let \(\mathcal{D}=\{(I^{(i)},\Pi^{(i)})\}_{i=1}^{N}\) denote an unordered set of $N$ posed input views, where $I^{(i)}\in\mathbb{R}^{H\times W\times 3}$ is the RGB
image of view $i$ and $\Pi^{(i)}:\mathbb{R}^{3}\rightarrow\mathbb{R}^{2}$ is its perspective projection. The goal of CanonicalGS is to convert these unordered observations into a stable latent world before GP decoding, so that reliable additional views consolidate scene evidence rather than introduce inconsistent or redundant primitives. At a high level, the set-to-scene mapping can be defined as
\begin{equation}
    \mathcal{Z} =
    \operatorname{Agg}\!\left(\mathcal{E}
    \right),\qquad \mathcal{E} = \left\{\operatorname{Ext}\left(I^{(i)},\Pi^{(i)}\right)\right\}_{i=1}^{N},
    \label{eq:method_overview}
\end{equation}
where $\operatorname{Ext}(\cdot)$ extracts per-view visual evidence, and $\operatorname{Agg}(\cdot)$ reprojects and aggregates this evidence in a shared scene space. The resulting latent scene is then decoded into a renderable GP set:
\begin{equation}
    \mathcal{G} = \operatorname{Dec}\left(\mathcal{Z}\right).
    \label{eq:decode_overview}
\end{equation}

\subsection{View-Centric Evidence Formation}
\vspace{-1mm}

In the first stage, for each input view, CanonicalGS estimates dense visual features, depth, and uncertainty before any scene-level aggregation is performed.

\textbf{Per-view feature extraction.}
Given the $i$-th input view $I^{(i)}$, we use a pretrained DINO-v2 backbone~\citep{oquab2023dinov2} to extract patch-level visual features. A lightweight upsampling decoder then fuses multi-scale features and restores them to the input image resolution~\citep{ranftl2021visiontransformersdenseprediction}, yielding an $L$-dimensional dense feature map
\(F^{(i)}_\mathrm{view} \in \mathbb{R}^{H\times W\times L}\). The feature extractor is applied independently to each view, avoiding introducing view-order dependence before observations are lifted into the shared scene space.

\textbf{Uncertainty-aware depth estimation.}
Depth provides the 2.5D bridge from images to scene space, while uncertainty determines which lifted observations should be trusted. We use a cost-volume depth module inspired by plane-sweep stereo~\citep{chen2024mvsplat,xu2023unifying,xu2025depthsplat}. Given $K$ sampled depth hypotheses $\{d_k\}_{k=1}^{K}$ and for each pixel location $m$ in the reference view $i$, the $k$-th hypothesis is back-projected to the scene space and reprojected to the source view $j$:
\begin{equation}
    m^{(i,j)}_k
    =
    \Pi^{(j)}\!\left(\left(\Pi^{(i)}\right)^{-1}(m,d_k)\right).
\end{equation}
The pairwise matching score is computed by feature correlation:
\begin{equation}
    S_k^{(i,j)}(m)
    =
    \frac{
    \left\langle F^{(i)}_\mathrm{view}(m), F^{(j)}_\mathrm{view}(m^{(i,j)}_k) \right\rangle
    }{\sqrt{L}}.
\end{equation}
For an $N$-view input set, the reference-view score volume is computed by averaging across all source-view scores:
\begin{equation}
    S_k^{(i)}(m)=\frac{1}{N-1}\sum_{j\neq i}S_k^{(i,j)}(m),
\end{equation}
Given $F^{(i)}_\mathrm{view} \in \mathbb{R}^{H\times W\times L}$ and $S^{(i)} \in \mathbb{R}^{H\times W\times K}$, a UNet-like module~\citep{10.1007/978-3-319-24574-4_28,xu2025depthsplat} outputs a discrete depth probability volume:
\begin{equation}
    P^{(i)}
    =
    \operatorname{softmax}
    \left(
    \operatorname{UNet}\left(F^{(i)}_\mathrm{view},S^{(i)}\right)
    \right)\in\mathbb{R}^{H\times W\times K}.   
\end{equation}
The depth map $D^{(i)}$ and positional uncertainty map $U^{(i)}_{\mathrm{pos}}$ are defined to have per-pixel values:
\begin{equation}
    D^{(i)}(m)=\sum_{k=1}^{K} P_k^{(i)}(m)\,d_k,
    \qquad
    U^{(i)}_\mathrm{pos}(m)
    =
    \sqrt{\sum_{k=1}^{K}
    P_k^{(i)}(m)\left(d_k-D^{(i)}(m)\right)^2},
    \label{eq:depth_uncertainty}
\end{equation}
where larger positional uncertainty indicates more ambiguous geometry, occlusion, or weaker multi-view agreement. In addition, we predict an appearance uncertainty map from semantic features:
\begin{equation}
    U^{(i)}_{\mathrm{app}}(m)=
    \operatorname{ReLU}
    \left(
    \operatorname{MLP}\left(F^{(i)}_\mathrm{view}(m)\right)
    \right),
    \label{eq:appearance_uncertainty}
\end{equation}
where $\operatorname{MLP}(\cdot)$ denotes a multilayer perceptron (MLP), with the same parameters shared across all spatial locations $m$. The two uncertainty sources are converted into a reliability map:
\begin{equation}
    R_{\mathrm{view}}^{(i)}
    =
    \exp\left(-U^{(i)}_{\mathrm{pos}}\right)
    \odot
    \exp\left(-U^{(i)}_{\mathrm{app}}\right),
    \label{eq:view_certainty}
\end{equation}
where $\odot$ denotes element-wise multiplication. As a result, an observation is highly reliable only when it is geometrically well supported and visually certain. The resulting view-centric evidence is $\mathcal{E} =\left\{F^{(i)}_\mathrm{view},D^{(i)},R_{\mathrm{view}}^{(i)}\right\}_{i=1}^N$.

\subsection{Scene-Centric Evidence Aggregation}
\vspace{-1mm}
In the second stage, CanonicalGS turns unordered view-centric evidence into two voxel fields in scene space: a feature field for scene content and a reliability field for accumulated support.

\textbf{Scene-space rasterization.}
Lifted image observations form an \textit{irregular} 3D point set whose sampling density varies with depth, viewpoint, and occlusion. Aggregating this set directly would make the representation depend on the incidental sampling pattern of the input views. We therefore rasterize the lifted observations onto a shared voxel lattice $\mathcal{V}$ with a fixed origin and a fixed grid resolution.

For a pixel location $m$ in view $i$, its lifted scene position is
\begin{equation}
    x^{(i)}(m)
    =
    \left(\Pi^{(i)}\right)^{-1}
    \!\left(m,D^{(i)}(m)\right),
    \label{eq:lift_evidence}
\end{equation}
which is assigned to its containing voxel. 

\textbf{Reliability-guided aggregation.}
Observations inside the same voxel may include corroborating measurements, duplicates, and inconsistent estimates caused by occlusion or depth ambiguity. To favor reliable and mutually consistent evidence, we first select the most certain observation in each nonempty voxel $v\in\mathcal{V}$ as a local representative:
\begin{equation}
    (i^\star,m^\star)
    =
    \operatorname*{arg\,max}_{(i,m)\in\Omega(v)} R_{\mathrm{view}}^{(i)}(m),    \label{eq:representative}
\end{equation}
where $\Omega(v)=\{(i,m)\mid x^{(i)}(m)\in v\}$ denotes the set of view-location index pairs, with lifted scene positions assigned to voxel $v$.
Each observation then receives a weight that combines its reliability with nonnegative feature agreement to this representative:
\begin{equation}
    W^{(i)}(m,v)
    =
    R_{\mathrm{view}}^{(i)}(m)
    \left[
    \frac{
    \left\langle F^{(i)}_\mathrm{view}(m), F^{(i^\star)}_\mathrm{view}(m^\star)\right\rangle
    }{
    \|F^{(i)}_\mathrm{view}(m)\|\,\|F^{(i^\star)}_\mathrm{view}(m^\star)\|
    }
    \right]_+,
    \qquad
    (i,m)\in\Omega(v),
    \label{eq:aggregation_weight}
\end{equation}
where $[\cdot]_+=\max(\cdot,0)$ and $W^{(i)}(m,v)$ denotes the reliability-guided aggregation weight for location $m$ in view $i$ assigned to voxel $v$. Thus, an observation contributes strongly only when it is both reliable and consistent with the local representative. The aggregated scene reliability and feature fields are
\begin{equation}
   R_{\mathrm{scene}}(v)
    =  \sum_{(i,m)\in\Omega(v)}
    W^{(i)}(m,v),
    \qquad
    {F}_\mathrm{scene}(v)
    =
    \frac{
    \sum_{(i,m)\in\Omega(v)}
    W^{(i)}(m,v)F^{(i)}_\mathrm{view}(m)
    }{
    R_{\mathrm{scene}}(v)+\epsilon
    },
    \label{eq:canonical_fields}
\end{equation}
where $\epsilon >0$ is a small constant to avoid potential division by zero. The scene reliability field records accumulated support, while the feature field stores the corresponding reliability-weighted scene descriptor. As more valid views observe the same scene region, $R_{\mathrm{scene}}(v)$ increases and ${F}_\mathrm{scene}(v)$ becomes dominated by mutually consistent evidence. This instantiates the scene representation in Eq.~\eqref{eq:method_overview} as $\mathcal{Z}=\{{F}_{\mathrm{scene}},R_{\mathrm{scene}}\}$. The same feature field can also support downstream perception by splatting latent features into a target view and applying a task-specific head~\citep{wewer24latentsplat}.

\textbf{GP decoding.}
In the final stage, the decoder converts the scene fields into a renderable GP set. Let
\begin{equation}
    \mathcal{G}=\{\mu(v),\Sigma(v),\alpha(v),h(v)\}_{v\in\mathcal{V}},   \label{eq:dec_instantiation}
\end{equation}
 where each primitive consists of a mean $\mu(v)$, covariance $\Sigma(v)$, opacity $\alpha(v)$, and spherical harmonic appearance coefficients $h(v)$. We first compute the opacity by bounding the accumulated scene reliability and then apply an MLP head:
\begin{equation}
    r(v)=1-\exp(-R_{\mathrm{scene}}(v)),
    \qquad
    \alpha(v)=\phi(r(v)) = \operatorname{Sigmoid}\left(\operatorname{MLP}\left(r(v)\right)\right),
    \qquad
    \frac{\partial \phi(r)}{\partial r}\ge 0.
    \label{eq:constrained_opacity}
\end{equation}
The nondecreasing constraint, implemented with nonnegative weights in the MLP head, encourages opacity to increase or remain saturated as reliable support accumulates. The remaining primitive attributes are decoded from the scene feature field:
\begin{equation}
    \delta(v),\ \Sigma(v),\ h(v)
    =
    \psi\!\left(F_{\mathrm{scene}}(v)\right),
    \label{eq:gp_decode}
\end{equation}
where $\delta(v)$ is a position offset. The mean is anchored at the representative lifted point selected in Eq.~\eqref{eq:representative} and refined by this offset:
\begin{equation}
    \mu(v)=x^{(i^\star)}(m^\star)+\delta(v),
    \qquad
    (i^\star,m^\star)\in\Omega(v).
\end{equation}
Thus, scene reliability controls opacity, and scene features determine geometry and appearance.
\section{Experiments}\label{sec:exp}
\vspace{-1mm}
In this section, we evaluate CanonicalGS from both rendering and representation perspectives, asking whether additional views become reliable scene evidence, whether the learned representation remains stable under changing input sets, and which design choices produce these effects.

\begin{table}[t]
\centering
\caption{\small Quantitative novel view synthesis results on RE10K~\citep{zhou2018stereo}. Dagger ($\dagger$) marks Gaussian-space merging variants that combine decoded primitives from the corresponding base model.}
\small
\setlength{\tabcolsep}{1.5pt}
\renewcommand{\arraystretch}{0.95}
\resizebox{\columnwidth}{!}{%
\begin{tabular}{lcccccccccccc}
\toprule
\multirow{2}{*}{{Method}} &
\multicolumn{3}{c}{{$2$ views}} &
\multicolumn{3}{c}{{$4$ views}} &
\multicolumn{3}{c}{{$6$ views}} &
\multicolumn{3}{c}{{$8$ views}} \\
\cmidrule(lr){2-4}\cmidrule(lr){5-7}\cmidrule(lr){8-10}\cmidrule(lr){11-13}
& PSNR$\uparrow$ & SSIM$\uparrow$ & LPIPS$\downarrow$
& PSNR$\uparrow$ & SSIM$\uparrow$ & LPIPS$\downarrow$
& PSNR$\uparrow$ & SSIM$\uparrow$ & LPIPS$\downarrow$
& PSNR$\uparrow$ & SSIM$\uparrow$ & LPIPS$\downarrow$ \\
\midrule
MVSplat
& 22.52 & 0.801 & 0.187
& 20.94 & 0.790 & 0.211
& 20.58 & 0.774 & 0.231
& 19.68 & 0.749 & 0.255 \\

DepthSplat
& 24.16 & 0.838 & 0.166
& 23.39 & 0.844 & 0.164
& 23.01 & 0.839 & 0.166
& 22.00 & 0.814 & 0.190 \\
\midrule
MVSplat$^{\dagger}$
& 18.78 & 0.623 & 0.422
& 19.10 & 0.631 & 0.415
& 19.09 & 0.626 & 0.420
& 18.92 & 0.622 & 0.418 \\

DepthSplat$^{\dagger}$
& 23.37 & 0.805 & 0.202
& 23.26 & 0.803 & 0.208
& 23.01 & 0.795 & 0.213
& 22.35 & 0.774 & 0.230 \\

FreeSplat
& 21.74 & 0.782 & 0.209
& 21.66 & 0.799 & 0.213
& 21.62 & 0.796 & 0.222
& 20.07 & 0.746 & 0.257 \\

ZPressor
& 21.38 & 0.761 & 0.220
& 21.83 & 0.780 & 0.216
& 22.91 & 0.803 & 0.204
& 22.75 & 0.800 & 0.206 \\
\midrule
CanonicalGS (Ours)
& \textbf{24.22} & \textbf{0.840} & \textbf{0.164}
& \textbf{24.70} & \textbf{0.853} & \textbf{0.154}
& \textbf{24.82} & \textbf{0.857} & \textbf{0.149}
& \textbf{25.22} & \textbf{0.861} & \textbf{0.145} \\
\bottomrule
\end{tabular}%
}
\label{tab:free-re10k}
\end{table}

\subsection{Experimental Setups}
\vspace{-1mm}
\label{sec:setup}
\textbf{Datasets.}
We experiment on both indoor and outdoor scene collections.
For indoor scenes, we train and evaluate on RealEstate10K~\citep[RE10K,][]{zhou2018stereo}, which provides video sequences with camera poses and diverse room-scale motion.
For outdoor scenes, we fine-tune and evaluate on DL3DV~\citep{ling2024dl3dv}, which contains larger camera baselines and more complex appearance variation.
For representation stability, we use the RE10K test split and obtain semantic pseudo-labels with Mask2Former~\citep{cheng2021mask2former}, allowing us to probe whether the learned latent scene also supports downstream perception.

\textbf{Implementation details.}
All input images are resized to $256\times256$.
We train CanonicalGS in two stages.
First, we pretrain the depth-related modules, including the ViT backbone, upsampling head, and UNet, by distilling Depth Anything V2 predictions~\citep{depth_anything_v2}, using a learning rate of $10^{-4}$.
Because monocular teacher depths have arbitrary scale and shift, we follow the affine-invariant loss of~\cite{ranftl2020towards}: teacher and prediction depth maps are robustly centered by their medians, scaled by their mean absolute deviations, and then compared with an absolute error.
Second, we fine-tune the full model end-to-end with differentiable rendering supervision, using mean squared error and LPIPS with $\lambda_{\mathrm{LPIPS}}=0.05$, following DepthSplat and PixelSplat~\citep{xu2025depthsplat,charatan2024pixelsplat}. We train on RE10K for 300{,}000 steps with a batch size of 2 on four NVIDIA RTX A6000 GPUs, and fine-tune the RE10K model on DL3DV for 100{,}000 steps.
We use AdamW~\citep{adamW}, with an initial learning rate of $10^{-6}$ for the ViT backbone, upsampling head, and UNet, and $10^{-4}$ for the remaining parameters.
The learning rate follows cosine annealing with $2,000$ warm-up steps and a minimum value of $10^{-8}$.
Unless otherwise stated, training uses two input views; this keeps the training protocol sparse and makes test-time view scalability a property of the architecture rather than a consequence of matching the test view count.
We use default volume resolutions of $[1024, 1024, 512]$ for RE10K and $[768, 768, 384]$ for DL3DV.

\begin{table}[t!]
\centering
\caption{\small Quantitative novel view synthesis results on DL3DV~\citep{ling2024dl3dv}.}
\small
\setlength{\tabcolsep}{1.5pt}
\renewcommand{\arraystretch}{0.95}
\resizebox{\columnwidth}{!}{%
\begin{tabular}{ l c c c c c c c c c c c c
}
\toprule
\multirow{2}{*}{{Method}} &
\multicolumn{3}{c}{{$2$ views}} &
\multicolumn{3}{c}{{$4$ views}} &
\multicolumn{3}{c}{{$6$ views}} &
\multicolumn{3}{c}{{$8$ views}}\\
\cmidrule(lr){2-4}\cmidrule(lr){5-7}\cmidrule(lr){8-10}\cmidrule(lr){11-13}
& {\makecell{PSNR$\uparrow$}} & {\makecell{SSIM$\uparrow$}} & {\makecell{LPIPS$\downarrow$}} & {\makecell{PSNR$\uparrow$}} & {\makecell{SSIM$\uparrow$}}& {\makecell{LPIPS$\downarrow$}} & {\makecell{PSNR$\uparrow$}} & {\makecell{SSIM$\uparrow$}}& {\makecell{LPIPS$\downarrow$}} & {\makecell{PSNR$\uparrow$}} & {\makecell{SSIM$\uparrow$}}& {\makecell{LPIPS$\downarrow$}} \\
\midrule
MVSplat
& 17.35 & 0.507 & 0.416 & 17.20 & 0.535      
& 0.385 & 17.57 & 0.532 & 0.395 & 17.12 & 0.506 & 0.426 \\

DepthSplat & 19.00 & \textbf{0.596} & 0.331 & 19.60 & \textbf{0.670}      
& 0.287 & 19.14 & 0.667 & 0.283 & 17.97 & 0.626 & 0.320 \\
\midrule
MVSplat$^{\dagger}$
& 15.24 & 0.362 & 0.633 & 15.14 & 0.367      
& 0.621 & 15.65 & 0.379 & 0.631 & 15.37 & 0.372 & 0.639 \\

DepthSplat$^{\dagger}$
& 18.45 & 0.562 & 0.385 & 18.83 & 0.614      
& 0.359 & 18.52 & 0.603 & 0.366 & 17.71 & 0.570 & 0.396 \\

FreeSplat
& 16.69 & 0.463 & 0.443 & 17.93 & 0.597      
& 0.348 & 18.78 & 0.620 & 0.345 & 18.20 & 0.590 & 0.372 \\

ZPressor
& 16.27 & 0.440 & 0.453 & 16.82 & 0.501      
& 0.414 & 17.96 & 0.540 & 0.395 & 17.74 & 0.525 & 0.409 \\
\midrule
{CanonicalGS (Ours)}
& \textbf{19.26} & 0.592 & \textbf{0.331} & \textbf{19.78} & 0.665 & \textbf{0.284} & \textbf{20.05} & \textbf{0.667} & \textbf{0.278} & \textbf{20.21} & \textbf{0.676} & \textbf{0.269} \\
\bottomrule
\end{tabular}
}

\label{tab:free-dl3dv}
\end{table}

\begin{figure*}[t!]
	\centering
	\includegraphics[width=1.0\linewidth]{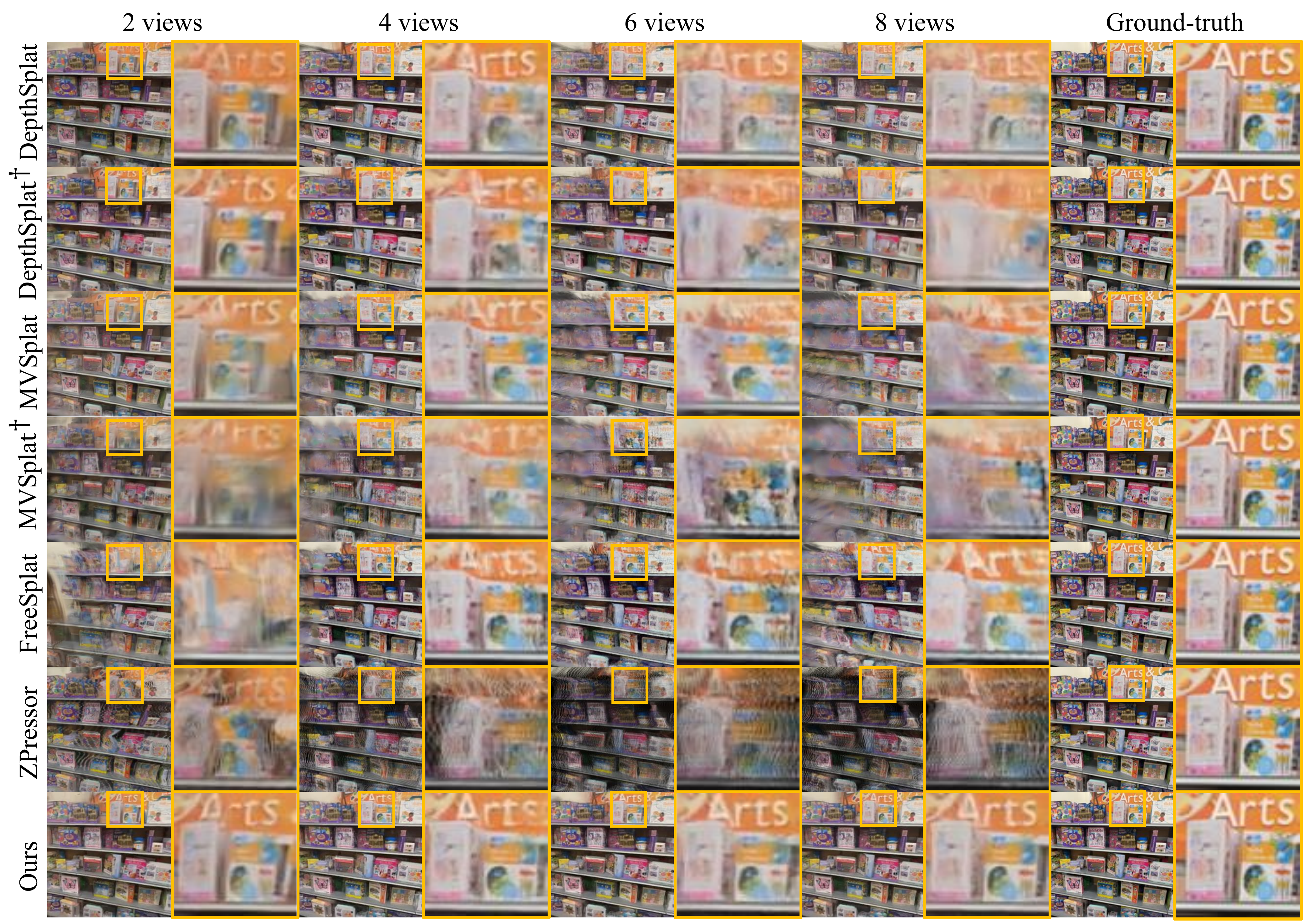}	
	\caption{\small Qualitative novel view synthesis results on DL3DV with increasing numbers of input views. Yellow boxes highlight regions where CanonicalGS benefits from additional context. Best viewed zoomed in.}
	\label{fig:vis1}
\end{figure*}

\subsection{Main Results}

\textbf{Novel view synthesis.} We first evaluate whether CanonicalGS can improve rendering quality as more input views are provided. 
During evaluation, we vary the number of input views from $2$ to $8$ and render $50$ target views for each scene. 
To stress the expressiveness of the learned representation, we set the target-view span to be 20\% larger than the input-view span. 
This protocol is substantially more challenging than those used in DepthSplat~\citep{xu2025depthsplat}, FreeSplat~\citep{wang2024freesplat}, and ZPressor~\citep{wang2025zpressor}, which typically evaluate $3$ to $8$ target views within the input span. We compare against three families of FFGS methods.
DepthSplat~\citep{xu2025depthsplat} and MVSplat~\citep{chen2024mvsplat} are pixel-aligned baselines.
FreeSplat~\citep{wang2024freesplat} and ZPressor~\citep{wang2025zpressor} represent latent- or feature-space merging approaches.
We also evaluate Gaussian-space merging variants by transplanting AnySplat-style merging~\citep{jiang2025anysplat} onto DepthSplat and MVSplat; these variants, denoted by $\dagger$, test whether post-decoding primitive merging alone can provide the scalability that CanonicalGS seeks before decoding.
We report PSNR, structural similarity \citep[SSIM,][]{1284395}, and learned perceptual image patch similarity~\citep[LPIPS,][]{8578166} values.

Tables~\ref{tab:free-re10k} and~\ref{tab:free-dl3dv} show that CanonicalGS not only improves the overall rendering metrics, but also changes the way performance evolves with additional context.
Pixel-aligned methods tend to peak early or degrade as more views are supplied, because each new view contributes another set of view-centric hypotheses that may overlap or conflict after decoding.
Gaussian-space merging variants reduce primitive redundancy after GP prediction, but they still operate on already-decoded primitives, where it is difficult to distinguish corroborating evidence from inconsistent geometry and appearance estimates.
Latent merging methods improve flexibility, yet their aggregation is not explicitly tied to per-observation reliability in a shared scene coordinate system.
CanonicalGS instead aggregates uncertainty-weighted evidence before GP decoding, so extra views are more likely to refine the latent scene field rather than inject additional inconsistent primitives.

Quantitative metrics alone do not fully reveal the failure modes. 
We therefore provide qualitative comparisons in Fig.~\ref{fig:vis1}. 
As the number of input views increases, pixel-aligned methods often accumulate inconsistent color and geometry, producing overlay artifacts in regions where multiple view-centric predictions disagree.
Gaussian-space merging variants can reduce explicit primitive overlap, but their corrections occur after the ambiguous primitives have already been produced, so ghosting and local appearance shifts remain visible.
Feature-space merging methods are less tied to individual pixels, but they remain sensitive to depth and pose noise because the fusion space is not explicitly organized as uncertainty-weighted scene evidence.
In contrast, CanonicalGS maintains sharper, more stable renderings as input views increase because scene-centric aggregation consolidates rather than accumulates view-dependent evidence.

\subsection{Representation Stability Evaluation}
\label{sec:rep-stability}
Rendering metrics measure image fidelity, but they do not directly reveal whether a feed-forward representation becomes more stable as the input set grows. We therefore evaluate CanonicalGS from two complementary perspectives. First, for each scene we splat feature maps from increasing input-view sets and compute cosine similarity to the feature map rendered from the largest input set (\ie, $12$ views), which serves as the scene-specific reference. Second, we freeze the splatted features and train a linear semantic probe. We report pixel-wise accuracy over $150$ classes and frequency-weighted intersection over union (FWIoU).

\begin{figure}[t]
\centering
\begin{minipage}[t]{0.48\linewidth}
\centering
    \includegraphics[width=0.98\linewidth]{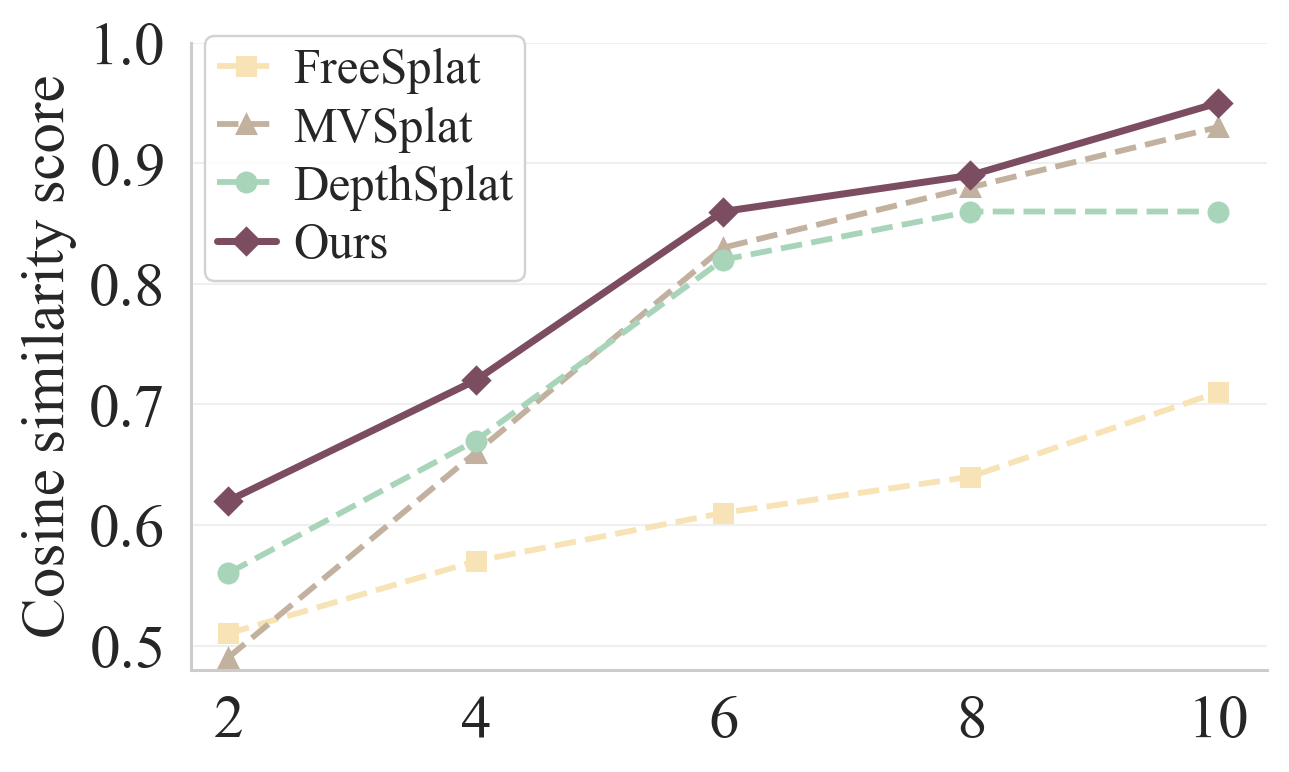}
{\small (a) Feature stability to the $12$-view reference}
\end{minipage}
\hfill
\begin{minipage}[t]{0.48\linewidth}
\centering
\includegraphics[width=0.98\linewidth]{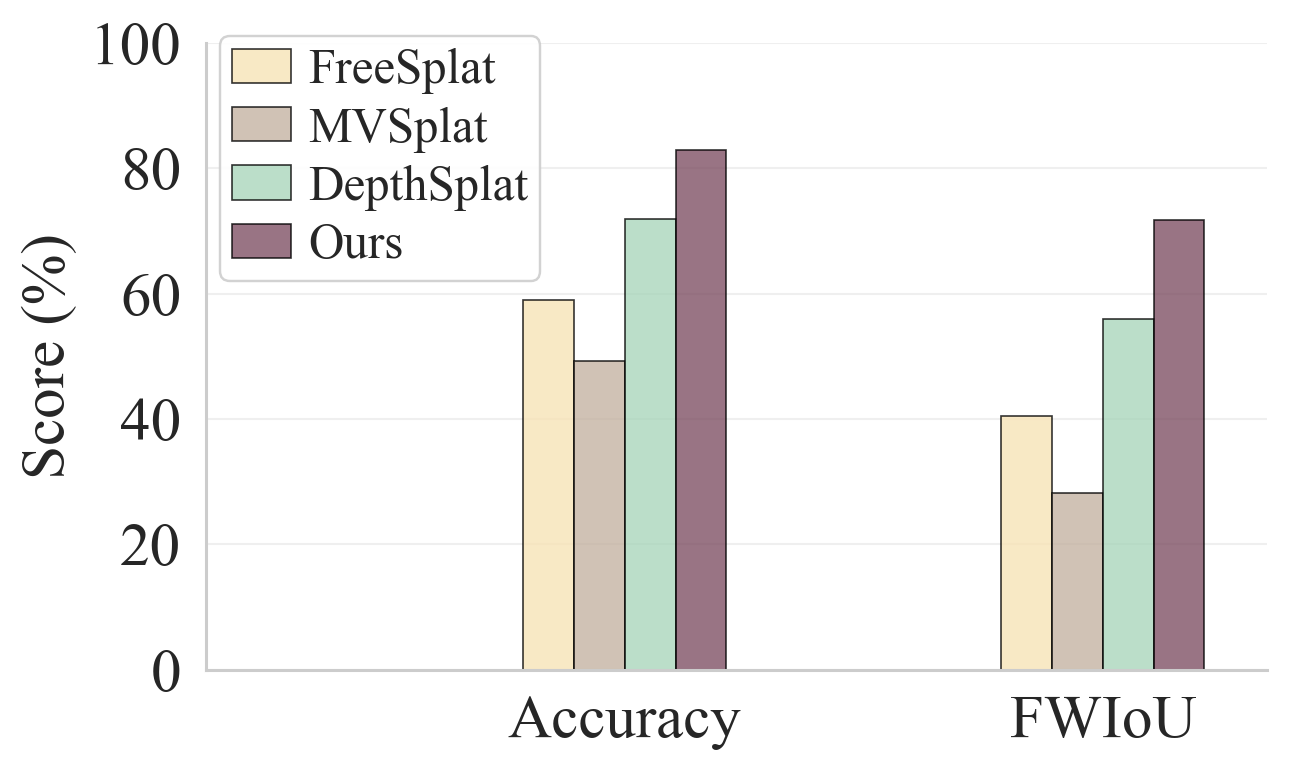}
{\small (b) Semantic segmentation}
\end{minipage}
\caption{\small Representation stability evaluation. Left: cosine similarity to the $12$-view reference under increasing input views, where higher curves indicate more stable features. Right: linear-probe semantic segmentation performance from splatted scene features.}
\label{fig:rep-stability}
\end{figure}

Figure~\ref{fig:rep-stability} shows that CanonicalGS is both more stable and more useful for downstream perception. In the left panel, its features move steadily toward the $12$-view reference as additional observations are added, whereas competing representations are more sensitive to view-set changes. This supports the role of scene-centric aggregation: new views are absorbed as reliable scene evidence rather than as independent view-conditioned hypotheses. The right panel further shows that these stabilized features carry stronger semantic information, yielding better linear-probe segmentation performance.

\begin{figure}[t]
\centering
\includegraphics[width=0.98\linewidth]{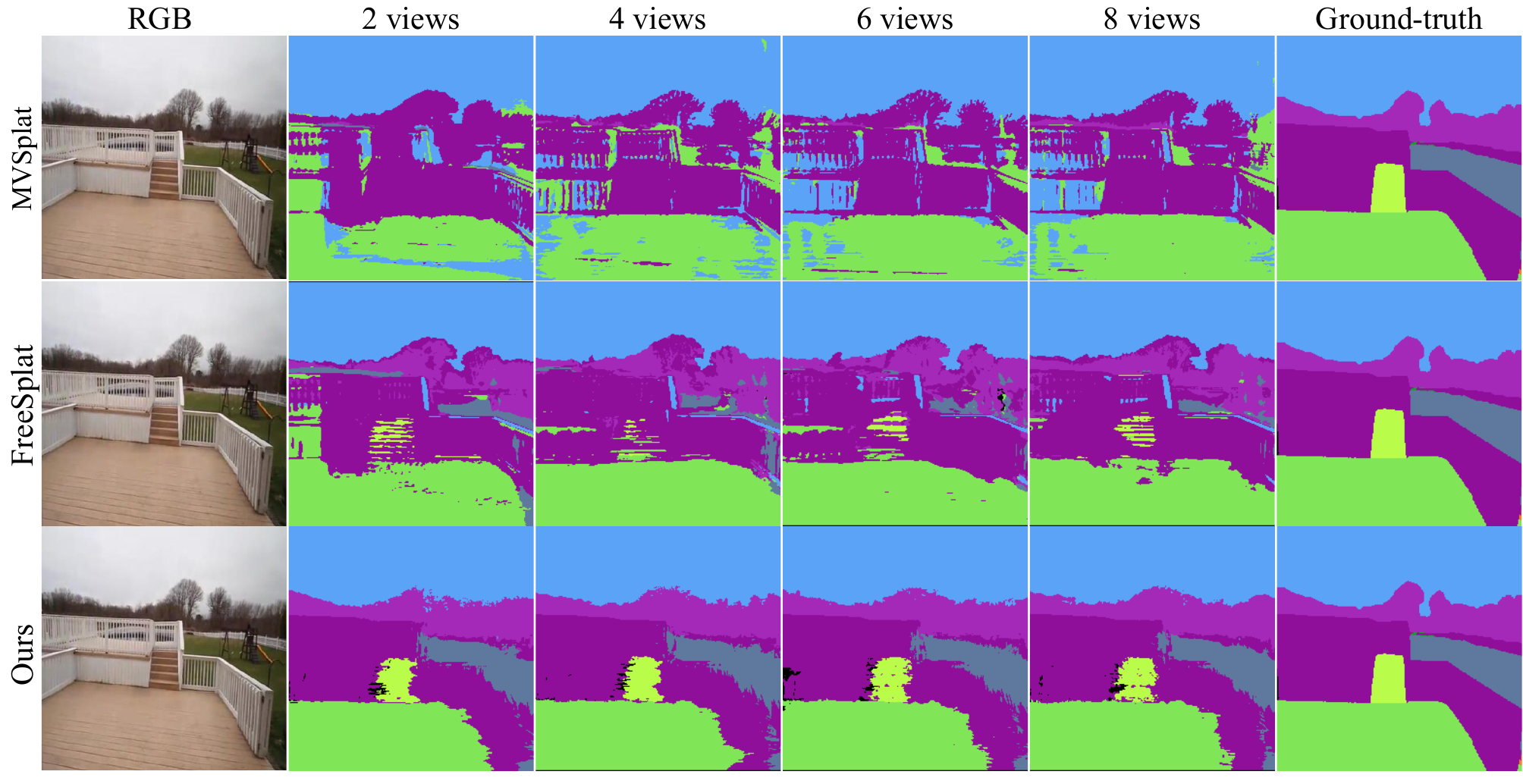}
\caption{\small Qualitative semantic segmentation from splatted scene features. CanonicalGS yields cleaner and more spatially coherent predictions, indicating that reliability-guided scene-centric aggregation preserves semantic structure in the decoded representation.}
\label{fig:seg-vis}
\end{figure}

Figure~\ref{fig:seg-vis} visualizes the same trend qualitatively. Baseline features often inherit local inconsistencies from view-dependent reconstruction artifacts, leading to fragmented semantic regions. CanonicalGS yields cleaner masks because uncertain observations are down-weighted before decoding while mutually consistent evidence is consolidated in a shared scene representation.

\subsection{Ablation Studies}
\label{sec:ablation}
We ablate CanonicalGS to separate the effects of scene-centric aggregation, reliability- and consistency-guided weighting, decoding constraints, training-view scale, and robustness to noisy depth estimates. The component and view-count ablations use DL3DV, while the noise robustness study uses RE10K. The reference model is the full CanonicalGS, and each variant changes one design choice while leaving the remaining pipeline unchanged.

\textbf{Scene-centric aggregation and decoding.}
The first three variants in Table~\ref{tab:ab_css} modify how lifted observations are consolidated before GP decoding. The w/o reliability variant keeps feature-similarity weighting but removes the reliability factor in Eq.~\eqref{eq:aggregation_weight}, so uncertain observations are no longer explicitly down-weighted. The w/o similarity variant keeps reliability weighting but removes agreement with the local representative, making duplicate or inconsistent observations easier to mix within a voxel. Average merging is a stronger simplification: it replaces the entire weighted aggregation rule with simple averaging. Plain decoding tests a different part of the pipeline: it keeps the aggregated scene fields but removes the reliability-conditioned decoding constraint. Table~\ref{tab:ab_css} shows that each component contributes to the final model, with the largest degradation when weighted aggregation is replaced by average merging. These results suggest that CanonicalGS benefits not only from placing observations in a shared scene space, but also from controlling which observations dominate that space and how accumulated support is used during decoding.

\textbf{Number of training views.}
The next ablation varies the number of input views used during training while keeping the full architecture and objective fixed. This experiment is complementary to Table~\ref{tab:ab_css}: rather than removing model components, it tests whether CanonicalGS can exploit denser multi-view evidence when available during learning. Table~\ref{tab:ab_views} shows a consistent improvement as the training view count increases, indicating that the model is not tied to a fixed sparse-view regime. Exposure to larger input sets helps CanonicalGS learn how to consolidate additional observations into the scene representation, which supports its view-scalable behavior.

\begin{table}[t]
\centering
\begin{minipage}[t]{0.52\linewidth}
\vspace{0pt}
\centering
\caption{\small Scene-centric aggregation and decoding ablation.}
\label{tab:ab_css}
\small
\setlength{\tabcolsep}{2.4pt}
\begin{tabular}{@{}lccc@{}}
\toprule
Variant & PSNR$\uparrow$ & SSIM$\uparrow$ & LPIPS$\downarrow$ \\
\midrule
Reference & 19.78 & 0.665 & 0.284 
\tabularnewline
\hline
w/o reliability & 18.65 & 0.592 & 0.314 \tabularnewline
w/o similarity & 19.20 & 0.623 & 0.298 \tabularnewline
Average merging & 18.58 & 0.590 & 0.306 \tabularnewline
Plain decoding & 18.70 & 0.594 & 0.311 \tabularnewline
\bottomrule
\end{tabular}
\end{minipage}
\hfill
\begin{minipage}[t]{0.42\linewidth}
\vspace{0pt}
\centering
\caption{\small Number of training views ablation.}
\label{tab:ab_views}
\small
\setlength{\tabcolsep}{4pt}
\begin{tabular}{@{}cccc@{}}
\toprule
\#Views & PSNR$\uparrow$ & SSIM$\uparrow$ & LPIPS$\downarrow$ \tabularnewline
\midrule
2 & 19.26 & 0.592 & 0.331 \tabularnewline
3 & 20.70 & 0.592 & 0.264 \tabularnewline
4 & 22.61 & 0.748 & 0.206 \tabularnewline
5 & 23.02 & 0.782 & 0.171 \tabularnewline
6 & 23.65 & 0.805 & 0.159 \tabularnewline
\bottomrule
\end{tabular}
\end{minipage}
\end{table}

\textbf{Noise robustness.}
Finally, we test whether the representation remains stable when the depth estimates used for lifting are corrupted. We perturb the estimated depth maps with zero-mean Gaussian noise at increasing normalized noise levels and evaluate on RE10K. Table~\ref{tab:noise_robustness} compares CanonicalGS with DepthSplat, a strong depth-based pixel-aligned baseline. Both models degrade as the perturbation increases, but CanonicalGS retains substantially higher PSNR and SSIM and incurs a smaller LPIPS increase. This gap suggests that reliability-guided scene-centric aggregation suppresses uncertain depth evidence before GP decoding, while direct depth-conditioned GP prediction is more exposed to local geometric perturbations.

\begin{table*}[t]
\centering
\caption{\small Noise robustness on RE10K under Gaussian noise perturbations to the estimated depth maps. $\Delta$ reports the relative change from each model's clean-depth baseline; for PSNR and SSIM, values closer to zero indicate smaller degradation, while for LPIPS, smaller positive increases indicate better robustness.}
\label{tab:noise_robustness}
\small
\setlength{\tabcolsep}{3.2pt}
\resizebox{\textwidth}{!}{%
\begin{tabular}{lcccccc@{\hspace{10pt}}cccccc}
\toprule
\multirow{2}{*}{Noise level} & \multicolumn{6}{c}{DepthSplat} & \multicolumn{6}{c}{CanonicalGS (Ours)} \\
\cmidrule(lr){2-7}\cmidrule(lr){8-13}
& PSNR$\uparrow$ & $\Delta$(\%) & SSIM$\uparrow$ & $\Delta$(\%) & LPIPS$\downarrow$ & $\Delta$(\%)
& PSNR$\uparrow$ & $\Delta$(\%) & SSIM$\uparrow$ & $\Delta$(\%) & LPIPS$\downarrow$ & $\Delta$(\%) \\
\midrule
Reference & 23.40 & -- & 0.841 & -- & 0.165 & -- & 24.91 & -- & 0.853 & -- & 0.151 & -- \\
0.05 & 23.31 & -0.4 & 0.832 & -1.1 & 0.185 & +12.1 & 24.88 & \textbf{-0.1} & 0.851 & \textbf{-0.2} & 0.155 & \textbf{+2.6} \\
0.10 & 23.06 & -1.5 & 0.818 & -2.7 & 0.210 & +27.2 & 24.81 & \textbf{-0.4} & 0.847 & \textbf{-0.7} & 0.162 & \textbf{+7.2} \\
0.25 & 22.22 & -5.0 & 0.776 & -7.7 & 0.273 & +65.5 & 24.39 & \textbf{-2.1} & 0.831 & \textbf{-2.5} & 0.182 & \textbf{+20.5} \\
0.50 & 21.03 & -10.1 & 0.720 & -14.4 & 0.345 & +99.1 & 23.50 & \textbf{-5.6} & 0.804 & \textbf{-5.7} & 0.209 & \textbf{+38.4} \\
\bottomrule
\end{tabular}%
}
\end{table*}

\section{Conclusion and Discussion}
We have introduced CanonicalGS, a feed-forward Gaussian splatting pipeline that organizes multi-view observations into a stable scene-centric representation before GP decoding. Rather than decoding view-centric primitives and reconciling them afterward, CanonicalGS first lifts visual, geometric, and reliability cues into a shared scene space, consolidates mutually consistent evidence, and then decodes a renderable GP set. Experiments show that this design improves novel view synthesis, produces features that are more stable as the input set grows, and transfers more effectively to downstream semantic segmentation. These results support the central premise of this work: additional views should strengthen a shared scene representation.

Despite these gains, CanonicalGS remains limited by the quality of its geometric inputs and by the structure of its scene representation. The current pipeline assumes reasonably accurate camera projections and depth estimates; pose error, depth-scale drift, heavy occlusion, and dynamic objects can place evidence in the wrong part of the canonical world before aggregation. A promising direction is joint canonicalization, where depth, pose, visibility, and scene evidence are refined together using the aggregated reliability field as feedback rather than treated as independent preprocessing cues. A second direction is richer uncertainty modeling. The current scalar reliability is effective, but calibrated occlusion likelihood and epistemic confidence could help distinguish unsupported regions from contradictory observations, and could further guide active view selection.

More broadly, CanonicalGS points toward feed-forward scene representations that are useful beyond rendering. Replacing the fixed voxel lattice with adaptive sparse or hierarchical scene fields would make our method more practical for room-, building-, and outdoor-scale scenes by allocating capacity to occupied or uncertain regions. Coupling the canonical world with language-level supervision could also turn splatted features into open-vocabulary maps for segmentation, 3D reasoning, and embodied planning. These directions are nontrivial because they require changing how evidence is formed, stored, and supervised, but they preserve the core idea of CanonicalGS: additional observations should increase reliable scene support rather than introduce view-conditioned clutter.
{
    \small
    \bibliographystyle{ieeenat_fullname}
    \bibliography{main}
}
\newpage
\appendix
\section*{Appendix}
This appendix provides experiments and visualizations that complement Secs.~\ref{sec:setup}--\ref{sec:ablation}. We first evaluate bounded-view rendering and zero-shot transfer, then analyze runtime and memory, and finally provide additional qualitative results for rendering, segmentation, and level-of-detail control.

\section{Bounded-View Evaluation}
We additionally evaluate the bounded-view protocol used by PixelSplat~\citep{charatan2024pixelsplat}, where target views lie within the input-view span. This setting is less extrapolative than the main evaluation in Sec.~\ref{sec:exp}, but it provides a useful comparison to prior sparse-view protocols. Table~\ref{tab:bounded} shows that CanonicalGS remains competitive in this easier setting, indicating that scene-centric aggregation improves rendering quality without relying on extrapolative target views.

\begin{table}[t]
\centering
\caption{\small Novel view synthesis results on RE10K~\citep{zhou2018stereo} in the bounded-view setting. Dagger ($\dagger$) marks Gaussian-space merging variants.}
\small
\setlength{\tabcolsep}{8pt}
\begin{tabular}{lccc}
\toprule
{Method} & PSNR$\uparrow$ & SSIM$\uparrow$ & LPIPS$\downarrow$ \\
\midrule
MVSplat & 26.39 & 0.869 & 0.128 \\
DepthSplat & 26.84 & 0.878 & 0.122 \\
MVSplat$^\dagger$ & 24.50 & 0.701 & 0.188 \\
DepthSplat$^\dagger$ & 25.22 & 0.840 & 0.166 \\
FreeSplat & 26.41 & 0.871 & 0.132 \\
ZPressor & 24.70 & 0.827 & 0.176 \\
\midrule
{CanonicalGS (Ours)} & \textbf{27.36} & \textbf{0.886} & \textbf{0.114} \\
\bottomrule
\end{tabular}
\label{tab:bounded}
\end{table}

\section{Zero-Shot Evaluation}
To test cross-dataset generalization, we train all models on RE10K with two input views and evaluate them on ACID~\citep{infinite_nature_2020} with four target views, following the DepthSplat split~\citep{xu2025depthsplat}. Table~\ref{tab:zero_acid} shows that CanonicalGS transfers best across datasets while using fewer parameters than FreeSplat~\citep{wang2024freesplat} and ZPressor~\citep{wang2025zpressor}. This result suggests that aggregating reliable evidence in scene space improves generalization, rather than merely increasing model capacity.

\begin{table}[t]
\centering
\caption{\small Zero-shot transfer to ACID~\citep{infinite_nature_2020}. All models are trained on RE10K with two input views and tested on ACID with four target views.}
\small
\setlength{\tabcolsep}{4.8pt}
\begin{tabular}{l S[table-format=2.2] S[table-format=1.3] S[table-format=1.3] c}
\toprule
{Method} & {PSNR$\uparrow$} & {SSIM$\uparrow$} & {LPIPS$\downarrow$} & {\#Params (M)} \\
\midrule
MVSplat & 22.75 & 0.834 & 0.178 & \textbf{12.0} \\
DepthSplat & 25.27 & 0.853 & 0.148 & 38.3 \\
MVSplat$^\dagger$ & 23.06 & 0.654 & 0.371 & 12.1 \\
DepthSplat$^\dagger$ & 24.98 & 0.748 & 0.261 & 38.3 \\
FreeSplat & 24.48 & 0.850 & 0.175 & 50.54 \\
ZPressor & 26.27 & 0.801 & 0.188 & 114.6 \\
\midrule
{CanonicalGS (Ours)} & \textbf{28.47} & \textbf{0.859} & \textbf{0.140} & 46.5 \\
\bottomrule
\end{tabular}
\label{tab:zero_acid}
\end{table}

\section{Runtime Analysis}
We report inference efficiency on DL3DV with $256\times256$ images, four input views, $50$ target views, and batch size one. Table~\ref{tab:runtime} compares rendering speed in frames per second (fps), peak GPU memory, average GP count, and rendering quality. CanonicalGS matches the highest fps, uses substantially less memory than pixel-aligned baselines, and keeps the decoded GP set compact while preserving the best PSNR. Compared with Gaussian-space merging, its efficiency comes from consolidating evidence before GP decoding rather than afterward pruning redundant primitives.

\begin{table}[t]
\centering
\caption{\small Runtime and memory analysis on DL3DV~\citep{ling2024dl3dv}. FPS reports rendered frames per second.}
\small
\setlength{\tabcolsep}{3.5pt}
\begin{tabular}{lcccc}
\toprule
{Method} & FPS$\uparrow$ & GPU (GB)$\downarrow$ & \#GPs (K)$\downarrow$ & PSNR$\uparrow$ \\
\midrule
MVSplat & 344.8 & 14.89 & 458.7 & 17.20 \\
DepthSplat & 303.0 & 14.83 & 262.1 & 19.60 \\
DepthSplat$^\dagger$ & 500.0 & 8.61 & 131.3 & 18.83 \\
FreeSplat & 555.6 & 8.85 & 187.9 & 17.93 \\
ZPressor & 400.0 & 9.00 & 393.2 & 16.82 \\
\midrule
{CanonicalGS (Ours)} & 555.6 & 9.98 & 172.3 & 19.78 \\
\bottomrule
\end{tabular}
\label{tab:runtime}
\end{table}

\section{Additional Qualitative Results}
\begin{figure*}[htbp]
\centering
\includegraphics[width=1.0\linewidth]{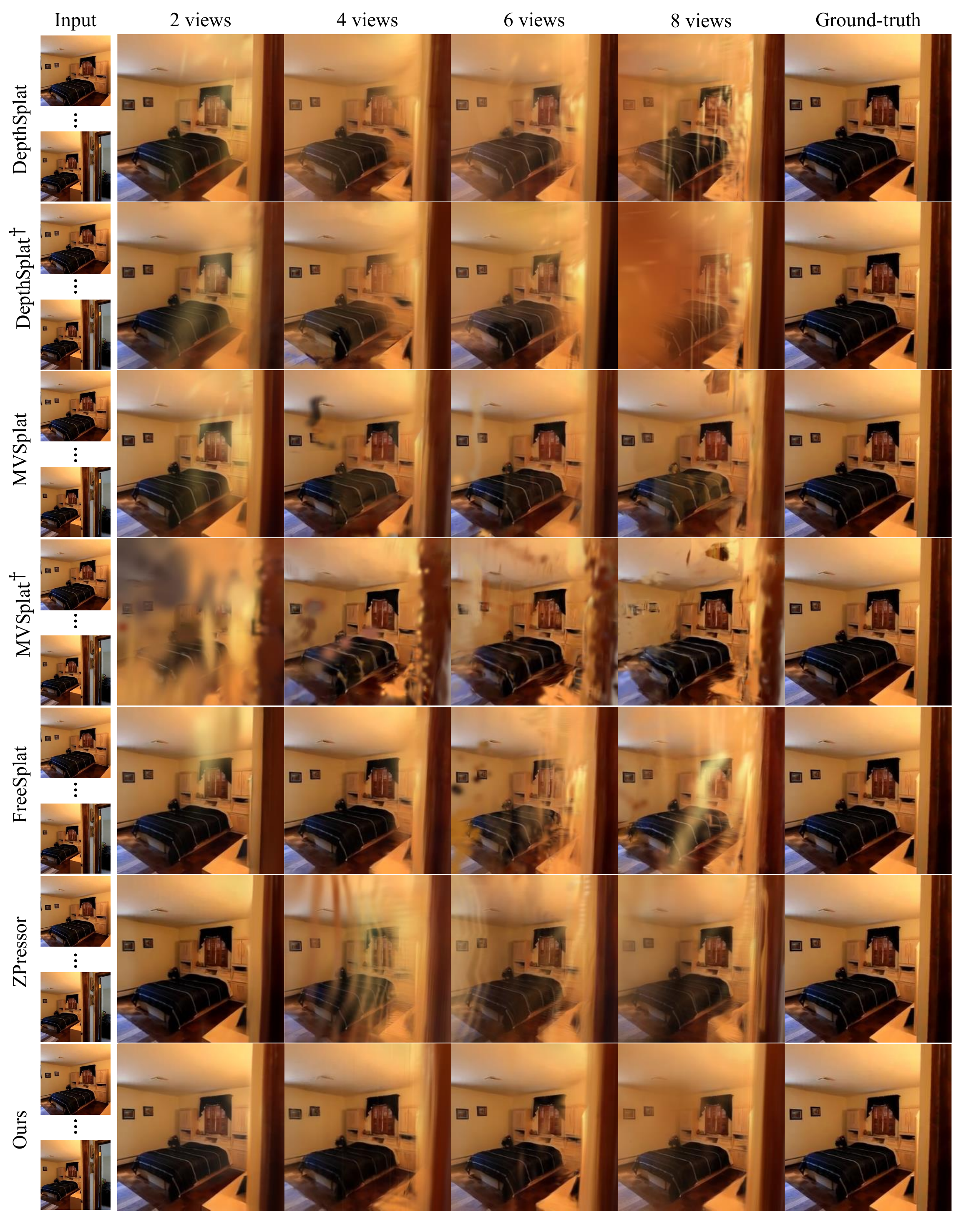}
\caption{\small Additional qualitative novel view synthesis comparisons on RE10K.}
\label{fig:re10k1}
\end{figure*}

\begin{figure*}[htbp]
\centering
\includegraphics[width=1.0\linewidth]{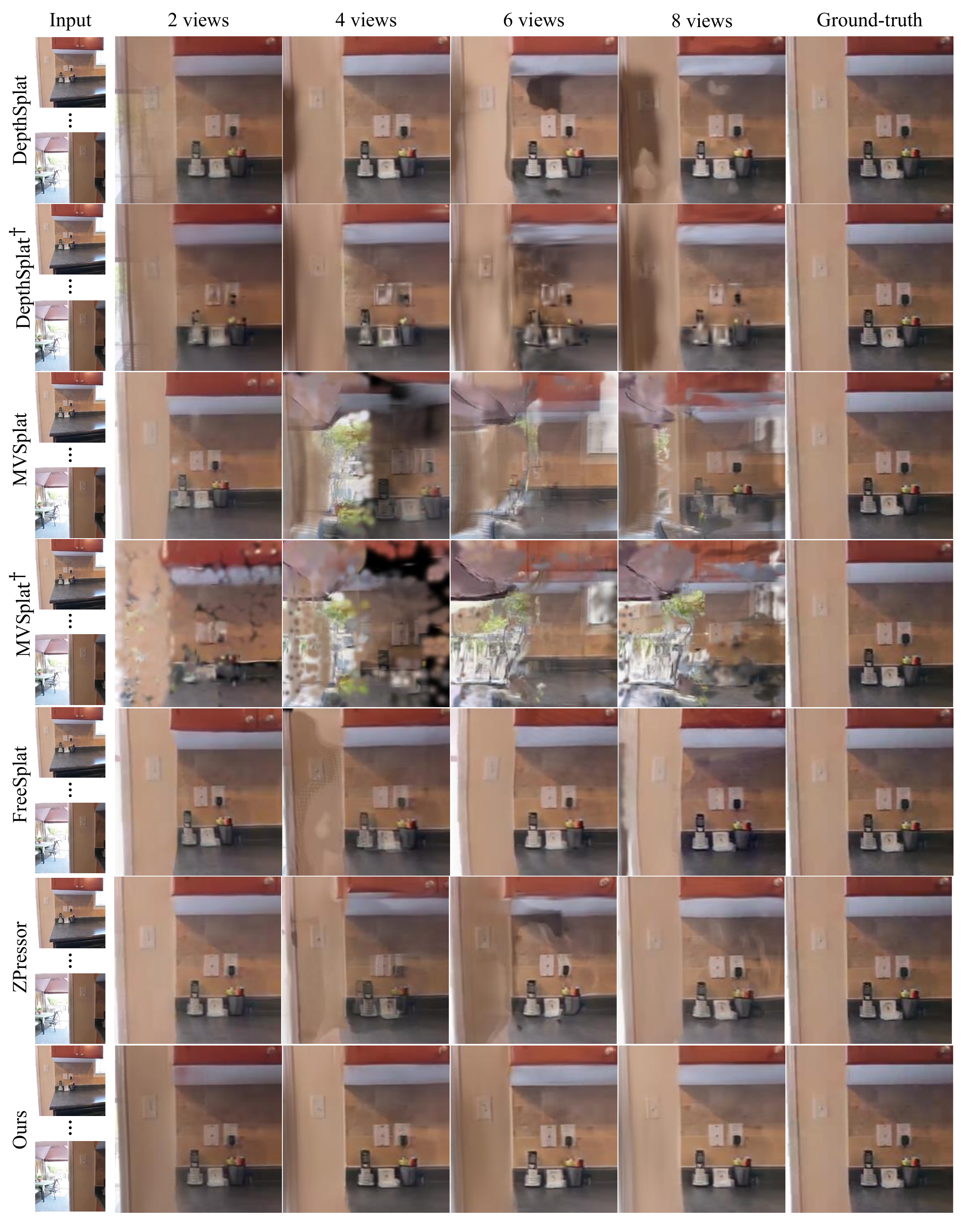}
\caption{\small Additional qualitative novel view synthesis comparisons on RE10K.}
\label{fig:re10k2}
\end{figure*}

\begin{figure*}[htbp]
\centering
\includegraphics[width=1.0\linewidth]{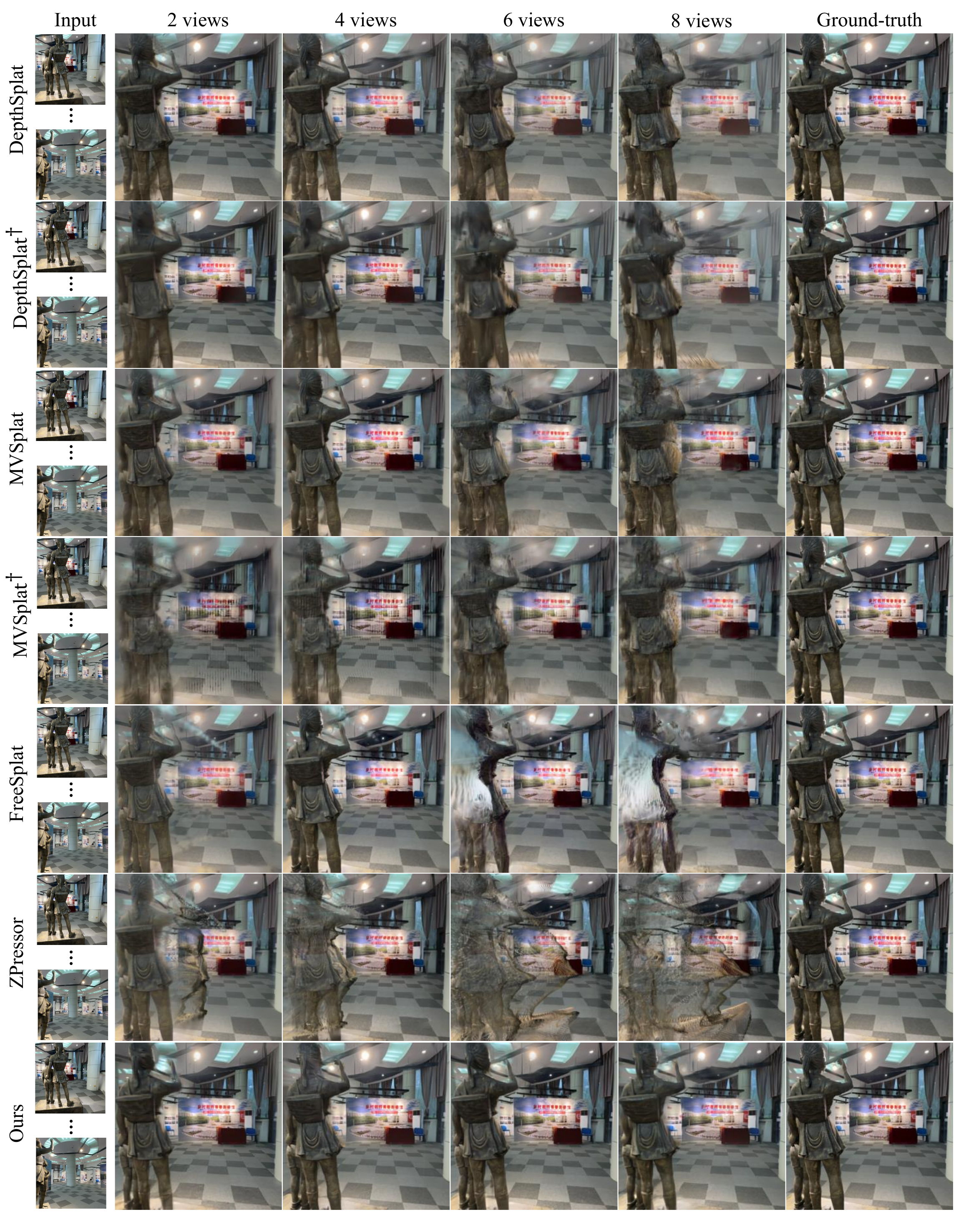}
\caption{\small Additional qualitative novel view synthesis comparisons on DL3DV.}
\label{fig:dl3dv}
\end{figure*}

Figs~\ref{fig:re10k1},~\ref{fig:re10k2}, and~\ref{fig:dl3dv} provide additional novel view synthesis comparisons on RE10K~\citep{zhou2018stereo} and DL3DV~\citep{ling2024dl3dv}. 
Across increasing input views, CanonicalGS more consistently preserves geometry and appearance, supporting the quantitative trend that additional observations are consolidated rather than accumulated as independent view-centric predictions.

\begin{figure*}[htbp]
\centering
\includegraphics[width=1.0\linewidth]{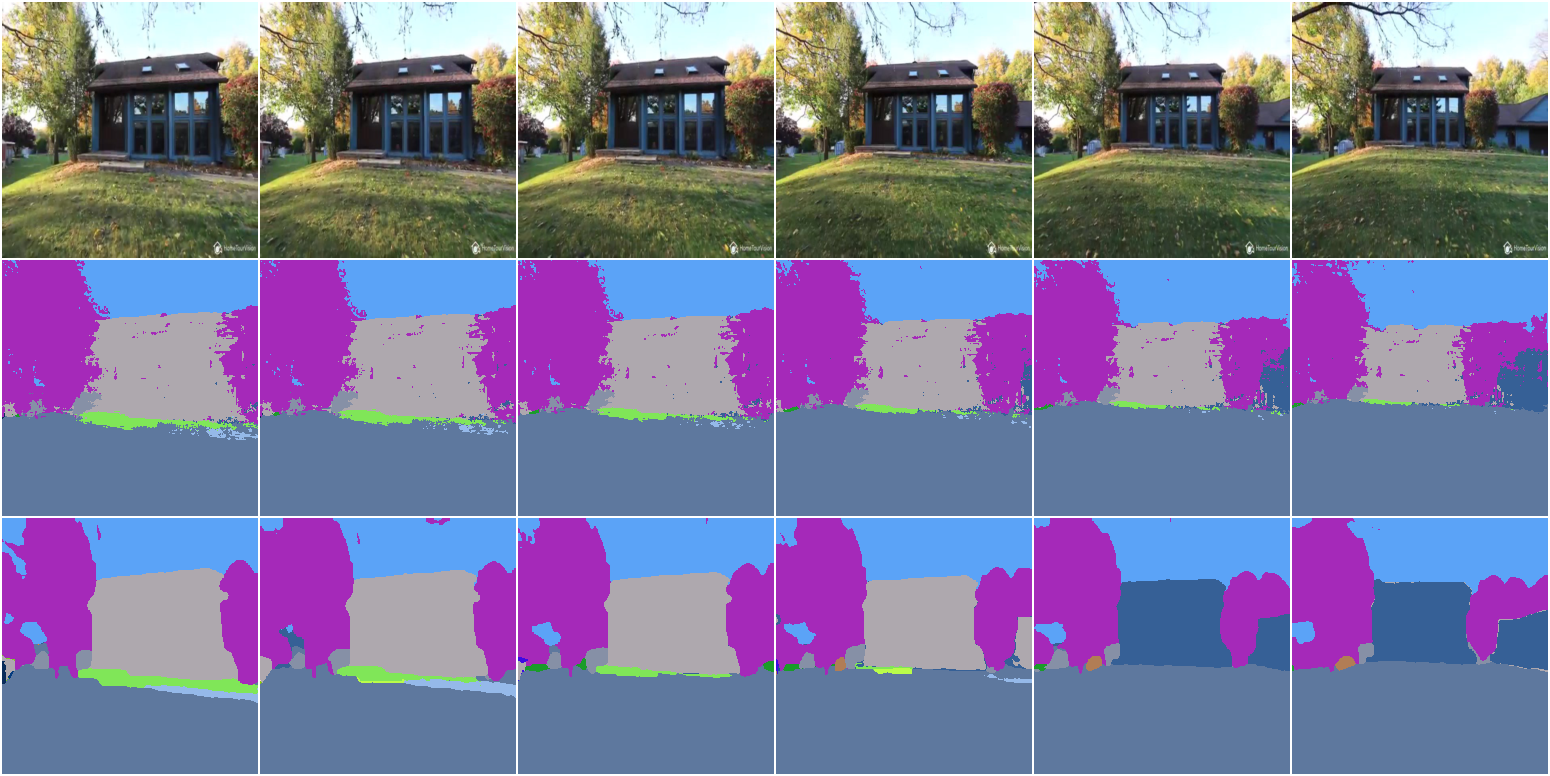}
\caption{\small Additional semantic segmentation visualization on RE10K. Rows show input images, linear-probe predictions from splatted CanonicalGS features, and ground-truth segmentation.}
\label{fig:seg1}
\end{figure*}

\begin{figure*}[htbp]
\centering
\includegraphics[width=1.0\linewidth]{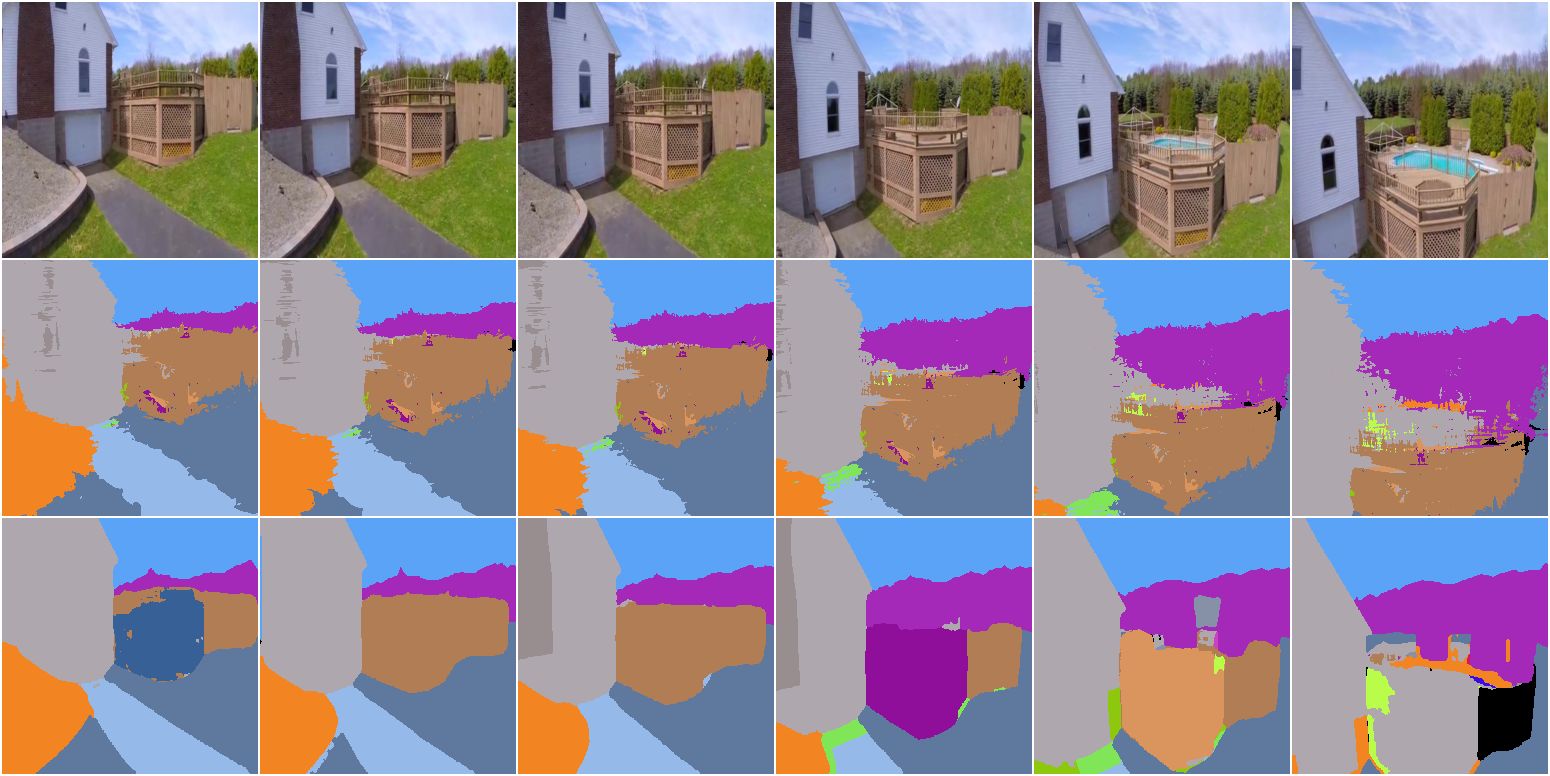}
\caption{\small Additional semantic segmentation visualization on RE10K.}
\label{fig:seg3}
\end{figure*}

\begin{figure*}[t!]
\centering
\includegraphics[width=1.0\linewidth]{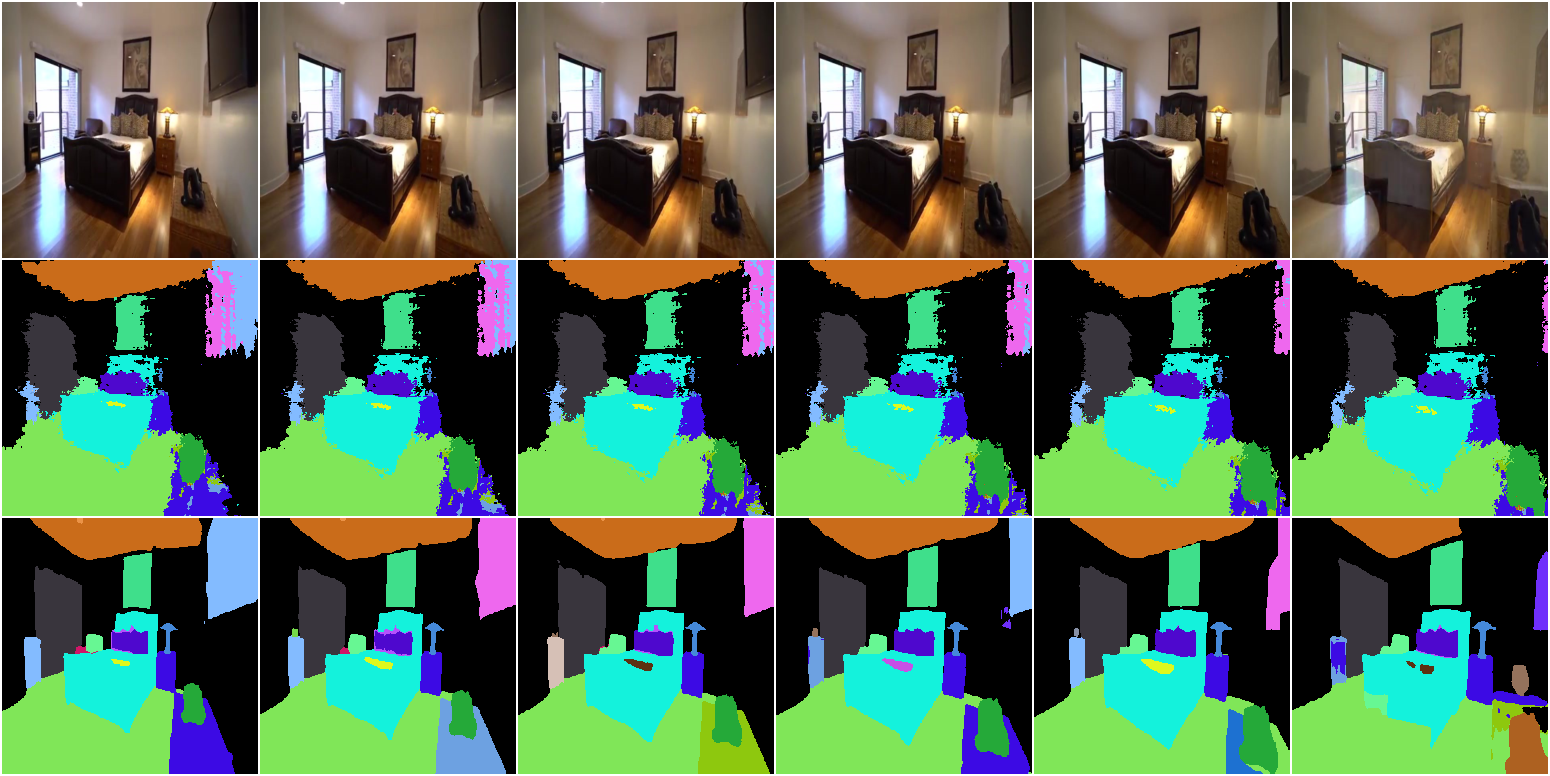}
\caption{\small Additional semantic segmentation visualization on RE10K.}
\label{fig:seg4}
\end{figure*}

Figs~\ref{fig:seg1}--\ref{fig:seg4} show additional semantic segmentation visualizations on RE10K. Each figure presents input images, linear-probe predictions from splatted CanonicalGS features, and ground-truth segmentation. The results show that scene-centric aggregation produces semantically coherent rendered features.

\section{Level-of-Detail Control}
\begin{figure*}[t!]
\centering
\includegraphics[width=1.0\linewidth]{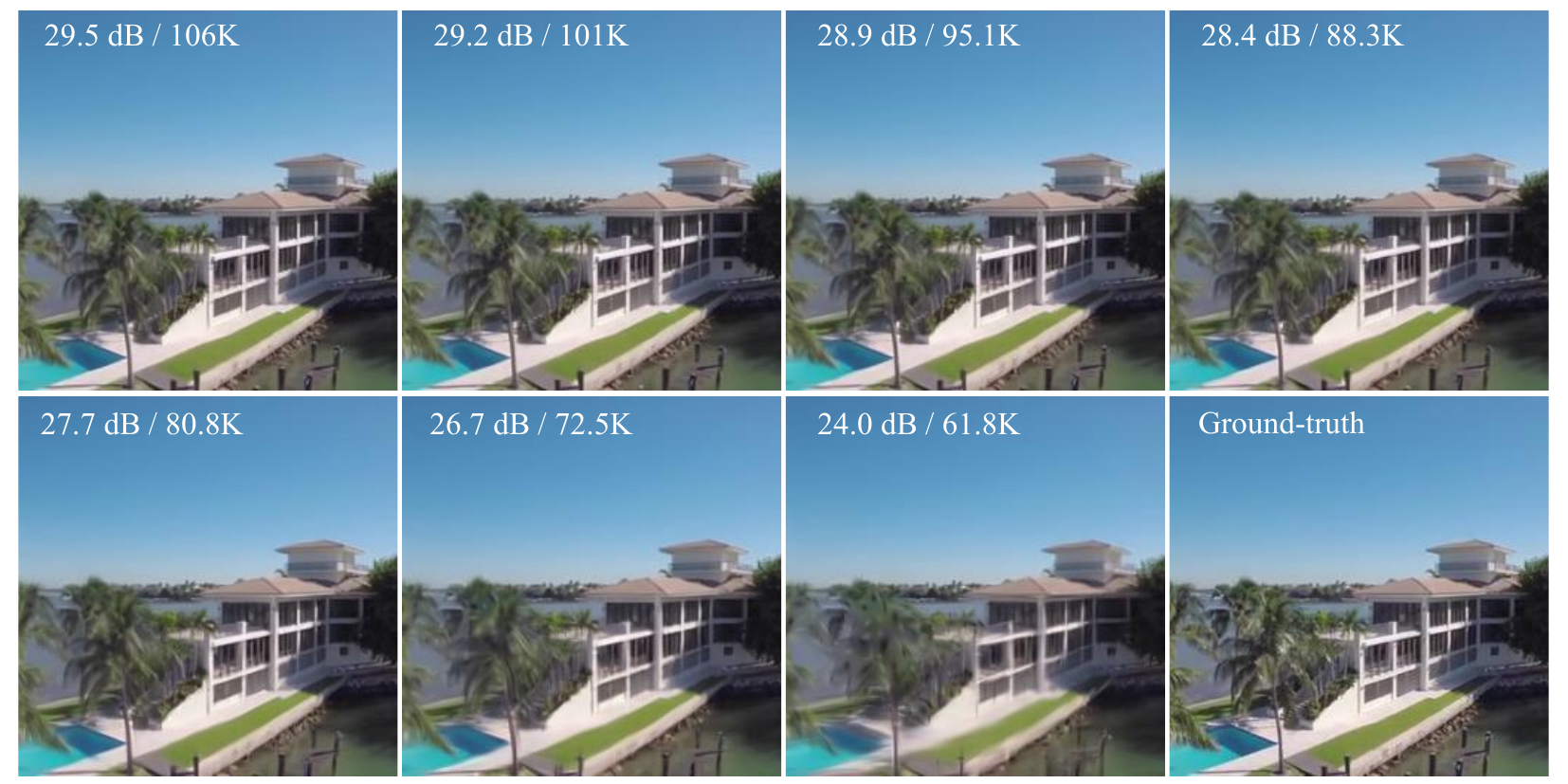}
\caption{\small Level-of-detail control by subsampling the scene-derived GP set. Each rendered example reports PSNR and the remaining number of GPs. CanonicalGS degrades smoothly under GP subsampling, showing a practical quality-compactness tradeoff without introducing hollow regions.}
\label{fig:lod}
\end{figure*}

Fig.~\ref{fig:lod} illustrates level-of-detail control by subsampling the scene-derived GP set. Because CanonicalGS decodes GPs from a scene-centric representation, reducing the number of primitives leads to a gradual quality change rather than the hollow artifacts often produced by removing view-aligned predictions. 
When more than roughly $70$K GPs are retained, the representation can be compressed with only modest quality degradation, suggesting a practical feed-forward compression route.

\clearpage
\newpage
\end{document}